\documentclass[sigconf]{acmart}

\usepackage[ruled,vlined]{algorithm2e}

\renewcommand\footnotetextcopyrightpermission[1]{}
\usepackage{graphicx}
\usepackage{url}
\usepackage{multirow}
\usepackage{comment}
\usepackage{pifont}
\usepackage{mathtools}
\usepackage{xcolor}
\usepackage{enumitem}
\usepackage{marvosym}
\usepackage{textcomp}
\usepackage{array}
\usepackage{booktabs}
\usepackage{multicol}
\usepackage{balance}
\usepackage{amsmath}
\usepackage{xcolor}
\usepackage{calc}
\usepackage{diagbox}
\usepackage{smile} 
\usepackage{xcolor,colortbl}
\usepackage{subfig}
\usepackage{amsfonts}
\usepackage{bbm}

\definecolor{darkpink}{rgb}{0.91, 0.33, 0.5}
\definecolor{sblue}{HTML}{2F6EBA}
\definecolor{sred}{HTML}{B02418}

\makeatletter
\newcommand{\thickhline}{%
    \noalign {\ifnum 0=`}\fi \hrule height 1pt
    \futurelet \reserved@a \@xhline
}

\usepackage{xspace}
\newcommand{\ours}{MINGLE\xspace}

\begin{document}

\title{Multimodal Fusion of EHR in \textit{Structures} and \textit{Semantics}: Integrating Clinical Records and Notes with Hypergraph and LLM}

\author{Hejie Cui}
\affiliation{%
  \institution{Emory University}
  \city{Atlanta}
  \state{GA}
  \country{USA}
}
\email{hejie.cui@emory.edu}
\author{Xinyu Fang}
\affiliation{%
  \institution{Zhejiang University}
  \city{Hangzhou}
  \state{Zhejiang}
  \country{China}
}
\email{xinyufang09@gmail.com}
\author{Ran Xu}
\affiliation{%
  \institution{Emory University}
  \streetaddress{400 Dowman Drive}
  \city{Atlanta}
  \state{Georgia}
  \country{USA}
  \postcode{30322}}
\email{ran.xu@emory.edu}
\author{Xuan Kan}
\affiliation{%
  \institution{Emory University}
  \streetaddress{400 Dowman Drive}
  \city{Atlanta}
  \state{Georgia}
  \country{USA}
  \postcode{30322}}
\email{xuan.kan@emory.edu}
\author{Joyce C. Ho}
\affiliation{%
  \institution{Emory University}
  \city{Atlanta}
  \state{GA}
  \country{USA}
}
\email{joyce.c.ho@emory.edu}
\author{Carl Yang}
\authornote{Carl Yang is the corresponding author.}
\affiliation{%
  \institution{Emory University}
  \city{Atlanta}
  \state{GA}
  \country{USA}
}
\email{j.carlyang@emory.edu}

\begin{abstract}
Electronic Health Records (EHRs) have become increasingly popular to support clinical decision-making and healthcare in recent decades. EHRs usually contain heterogeneous information, such as structural data in tabular form and unstructured data in textual notes. Different types of information in EHRs can complement each other and provide a more complete picture of the health status of a patient. While there has been a lot of research on representation learning of structured EHR data, the fusion of different types of EHR data (multimodal fusion) is not well studied. This is mostly because of the complex medical coding systems used and the noise and redundancy present in the written notes. In this work, we propose a new framework called \ours, which integrates both \textit{structures} and \textit{semantics} in EHR effectively. Our framework uses a two-level infusion strategy to combine medical concept semantics and clinical note semantics into hypergraph neural networks, which learn the complex interactions between different types of data to generate visit representations for downstream prediction. 
Experiment results on two EHR datasets, the public MIMIC-III and private CRADLE, show that \ours can effectively improve predictive performance by 11.83\% relatively, enhancing semantic integration as well as multimodal fusion for structural and textual EHR data. 
\vspace{-0.5ex}
\end{abstract}

\maketitle

\vspace{-1ex}
\section{Introduction}
\vspace{-0.3ex}

Electronic Health Records (EHRs) are widely used in healthcare and comprise heterogeneous data, including tabular records and unstructured clinical notes. Tabular records contain individual visits and are composed of a set of medical concepts such as diagnoses and medications. Clinical notes are long documents written by healthcare providers containing detailed information such as patient history, clinical findings, and laboratory test results.

Previous research has focused on structured EHR data modeling for predictive purposes~\cite{shickel2017deep}. The conventional approach involves modeling structured EHR data as numerical vectors, which are then processed using traditional machine learning (ML) models. However, this approach overlooks complex interactions between individual variables and does not consider interaction structures. To address this limitation, graph neural networks (GNNs) have been introduced to capture the hidden graph structure within EHR data~\cite{ochoa2022graph,choi2020learning}. More advanced approaches, such as hypergraph models, have been adopted to further capture higher-order interactions among visits and medical codes~\cite{xu2023hypergraph}. 
In this study, we aim to integrate structured EHR data with corresponding textual data. Although tabular and textual EHR integration has been studied quite recently~\cite{liu2022multimodal,li2022hi}, those studies mainly focused on pretraining and fine-tuning language models. We aim to combine structures and semantics and utilize the medical knowledge from the LLMs for improvement.
We focus on two types of textual information with EHR: medical code concept names and clinical notes. Integrating each type of information presents unique challenges due to the diversity of coding systems used, such as ICD-10, CPT, and SNOMED. This can cause discrepancies in mapping concept names across patient records and medical systems.
Clinical notes, on the other hand, are often riddled with errors, acronyms, and irrelevant or redundant information, creating challenges for integrating their semantics.

Integrating textual semantics from medical code concept names and clinical notes is crucial for ensuring accurate and comprehensive modeling of patient records. 
The recent success of large language models (LLMs) has created new opportunities in textual data integration. Inspired by this new trend, we explore LLMs to unify the modeling of medical concepts by generating semantic embeddings from LLM and fusing them with structural information of medical codes in the representation learning space. Furthermore, LLM can help identify and extract important semantics from clinical notes, which are then fused with patient visit representations to enhance the reasoning over visit-level interactions. 



\begin{figure}
	\centering
	\includegraphics[width=0.99\linewidth]{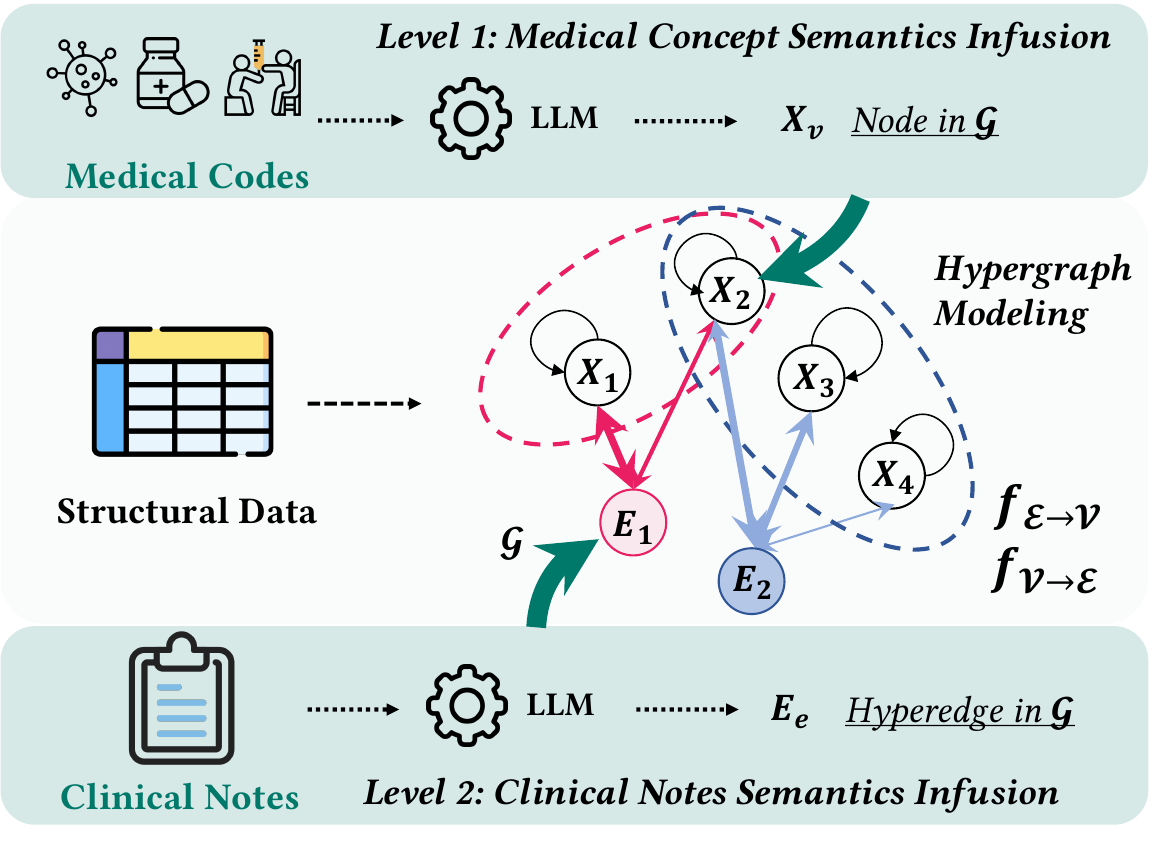}
 \vskip -0.5em
	\caption{The multimodal EHR fusion framework {\ours}.} 
 \vskip -0.5em
\label{fig:framework}
\vskip -0.3em
\end{figure}

We propose \ours, a multimodal EHR fusion framework that integrates structures and semantics from clinical records and notes, as shown in Figure~\ref{fig:framework}.  Our approach chooses the hypergraph neural network as the backbone, then further infuses \textit{medical concept semantics} and \textit{clinical notes semantics} into the structural modeling process using a two-level semantics infusion strategy and LLMs.

Experiment results and case studies on two EHR datasets demonstrate that \ours effectively enriches the representation of patient information. It jointly leverages the power of hypergraph GNNs to model complex relationships and harnesses the domain knowledge in LLMs and their strengths in natural language understanding. As a result, this integrated approach offers a comprehensive and nuanced analysis of EHR data, leading to more accurate and domain knowledge-enriched decision-making in healthcare.

\vspace{-1ex}
\section{Preliminaries}

\subsection{Structural and Textual Data in EHR} 
\label{sec:background}

EHR data is a digital collection of patient information, including structured clinical records and unstructured notes. The structured records are in a tabular format, where each row represents an individual patient visit, and columns are assigned to different medical codes. EHR also contains notes that complement the textual component. The combination of structured and textual data in EHR can provide a more comprehensive understanding of patient health and healthcare interactions.

\vspace{-1ex}
\subsection{Risk Prediction Problem}
Given a multimodal EHR dataset $\mathcal{D}=\left\{\mathcal{T}, \bf{\mathcal{N}}\right\}$, $\mathcal{T}=\{\mathcal{T}_p\}_{p=1}^{P}$ is the structured patient record that include $P$ rows of individual patient visits, and $\mathcal{N}_p$ is the corresponding clinical notes to each visit. 
The goal of our method is to train a predictive model that makes a clinical prediction for each given $p$-th visit $\mathcal{D}_p = \{\mathcal{T}_p, \mathcal{N}_p\}$.

\vspace{-1ex}
\subsection{Hypergraph Modeling of Structures in EHR}
Previous work on structured data modeling in EHR has demonstrated that transforming EHR tabular data to hypergraph can effectively encode the higher level co-occurrence relationships and interactions among visits and medical codes~\cite{xu2023hypergraph}, leading to effective visit representations for downstream prediction targets.

\noindent \textbf{Hypergraph Construction.} 
To transform tabular EHR data $\mathcal{T}$ into hypergraphs, each individual visit is modeled as a \textit{hyperedge} and each medical code as a \textit{node}. Each hyperedge connects all nodes that appeared in the corresponding visit. The hypergraph is denoted as $\cG = (\cV, \cE)$, where $\cV, \cE$ stand for \textit{nodes} and \textit{hyperedges} respectively. All \textit{nodes} belong to the set of medical codes $\cC$.

\noindent \textbf{Patient Visit Representation Learning.} 
We utilized the hypergraph neural network model from ~\citet{xu2023hypergraph}, which jointly learns node and hyperedge embeddings. In the $l$-th neural network layer, 
the update rule for node embedding is
\begin{equation}
\setlength{\abovedisplayskip}{0.1pt}
\setlength{\belowdisplayskip}{0.1pt}
\ \bX_{v}^{(l)} = f_{\mathcal{E} \rightarrow \mathcal{V}}\left(\cE_{v, \bE^{(l-1)}}\right),
\label{eq:node-update}
\end{equation}
and the rule for hyperedge embedding is 
\begin{equation}
\setlength{\abovedisplayskip}{0.1pt}
\setlength{\belowdisplayskip}{0.1pt}
\bE_{e}^{(l)} = f_{\mathcal{V} \rightarrow \mathcal{E}}\left(\cV_{e, \bX^{(l-1)}}\right),
\label{eq:hyperedge-update}
\end{equation}
where $\bX_{v}^{(l)}$ and $\bE_{e}^{(l)}$ denote the embeddings of node $v$ and hyperedge $e$ in the $l$-th layer ($1\leq l \leq L$), respectively. $\cE_{v, \bE}$ is the hidden representations of hyperedges that connect the node $v$, and $\cV_{e, \bX}$ is the hidden representations of nodes that are contained in the hyperedge $e$. 
The two message-passing functions $f_{\mathcal{V} \rightarrow \mathcal{E}}(\cdot)$ and $f_{\mathcal{E} \rightarrow \mathcal{V}}(\cdot)$ leverages multi-head self-attention, which help to identify important neighbors during propagation. Following the $L$ message passing layers, a $\operatorname{MLP}_{\text{CLS}}$ classification is used to convert the hyperedge embeddings to a value for binary classification,
\begin{equation}
\setlength{\abovedisplayskip}{0.1pt}
\setlength{\belowdisplayskip}{0.1pt}
\hat{y}_e = \sigma\left(\operatorname{MLP}_{\text{CLS}}\left(\|_{l=1}^{L}{\bE}_e^{(l)}\right)\right).
\label{eq:pred}
\end{equation}
The learning objective is defined as binary cross-entropy loss 
\begin{equation}
\setlength{\abovedisplayskip}{0.1pt}
\setlength{\belowdisplayskip}{0.1pt}
\ell_{\text{cls}} = -y\log(\hat{y}_e)-(1-y)\log(1-\hat{y}_e).
\label{eq:ce_loss}
\end{equation} 
\vskip -0.1em
\vspace{-0.8ex}
\section{Method}
\vspace{-0.1ex} 

Two textual semantics resources exist in the multimodal EHR dataset - the concept names of medical codes in tabular data and clinical notes. To infuse semantics into structural learning of hypergraph modeling, we propose a two-level strategy, as illustrated below.

\vspace{-1ex}
\subsection{Infusing \textit{Medical Concept Semantics} into the Structural Modeling of EHR data}
\label{sec:concept-semantics}

To reflect the structural contexts of nodes in the graph, we utilize the Deep Walk algorithm~\cite{perozzi2014deepwalk} to learn a structural latent representation $\bm s_{v}\in \mathbb{R}^{d_1}$ for each node $v$ in the hypergraph. This is particularly useful in the EHR modeling task as edges are sparse. To model the medical codes from different coding systems in a unified way, we first map the original code $v$ to the corresponding concept name $c_v$, then utilize GPT \texttt{text-embedding-ada-002} model to generate a semantic embedding $\bm c_{v}\in \mathbb{R}^{d_2}$, which contains clinical knowledge and context background from LLMs. 

We tried different ways to combine network-based and knowledge-based encoding, finding that simple concatenation achieves the best performance. Specifically, the $\bX_{v}^{(0)}$ in Eq.~(\ref{eq:node-update}) is initialized as the concatenation of both the structural feature $\bm{S}_{v}$ and the semantic feature $\bm{C}_{v}$ of the nodes in the hypergraph, 
\begin{equation}
\setlength{\abovedisplayskip}{0.1pt}
\setlength{\belowdisplayskip}{0.1pt}
\bm{X}_{v}^{(0)} = [\bm{S}_{v} ; \bm{C}_{v}].
\label{eq:infuse_concept}
\end{equation}
These fused node embeddings are utilized as the node feature initialization of the message-passing process, which induces the initial hyperedge embedding in Eq.~(\ref{eq:hyperedge-update}).

\vspace{-1ex}
\subsection{Infusing \textit{Clinical Note Semantics} into the Structural Modeling of EHR data}
\label{sec:note-semantics}

EHR datasets contain different types of clinical notes that serve specific purposes in documenting patient care. These notes include progress notes that track the patient's condition and treatment during hospitalization, nursing notes that provide information about the patient's daily care and response to treatment, radiology reports that interpret imaging results, and discharge summaries that give a comprehensive overview of the patient's hospital stay, including diagnoses, administered treatments, the patient's response to treatment, and follow-up care instructions. 
Out of all these types of notes, \textit{discharge summaries} are particularly valuable for integration with structured EHR data as they offer a granular summary that is instructive for continuous patient care.

In \ours, for each individual patient visit record $\mathcal{T}_p$ (correspond to a hyperedge $e \in \mathcal{E}$), we match the corresponding discharge summary $\mathcal{N}_p$ and filter irrelevant sections such as admission dates, services, etc. A document representation $\bm n_p$ is generated for each \textit{discharge summary} $\mathcal{N}_p$ with the GPT embedding model, resulting in a corpus semantic matrix $\bm{N}_e$ across all visits.
In order to further incorporate fine-grained semantics, we treat single nodes as additional hyperedges in the hypergraph by adding a self-loop on each node. The overall hyperedge semantics embeddings $\bm H_e$ is then the combination of the corpus semantic matrix $\bm{N}_e$ and the medical concept semantic matrix $\bm{C}_v$: 
\begin{equation}
\setlength{\abovedisplayskip}{0.1pt}
\setlength{\belowdisplayskip}{0.1pt}
    \bm{H}_{e} = \text{MLP}_1 \left(
    \begin{bmatrix}
        \bm{N}_e \\
        \bm{C}_v
    \end{bmatrix}
    \right).
\end{equation}
This leads to an enhancement of the central node semantics during its update from connected hyperedges, which also helps to establish a soft collaboration between fine-grained concept semantics and coarse-grained document semantics. 
Finally, we improve the hyperedge representation updating rule in Eq.~(\ref{eq:hyperedge-update}) as below:
\begin{equation}
\setlength{\abovedisplayskip}{0.1pt}
\setlength{\belowdisplayskip}{0.1pt}
\bm{E}_{e}^{(l)} = \text{MLP}_2([f_{\mathcal{V} \rightarrow \mathcal{E}}\left(\cV_{e, \bX^{(l-1)}}\right) ; \bm{H}_{e}]). 
\label{eq:infuse_note}
\end{equation}
This means that the hyperedge semantic embeddings $\bm H_e$ are incorporated into each message passing layer, along with the aggregated information from its connected nodes, to update the hyperedge representation.
\vspace{-1ex}
\section{Experiments}


\noindent \textbf{Datasets.} We have performed experiments on two clinical prediction datasets, MIMIC-III and CRADLE. The CRADLE dataset was collected from a large healthcare system in the United States. 
The MIMIC-III~\cite{johnson2016mimic} dataset contains 36,875 visits in all, represented by 7423 medical codes, with 12,353 visits being labeled. The CRADLE dataset contains 36,611 visits with 12,725 codes.
We divided them into a train, a validation, and a test set in the ratio of 7:1:2. As natural notes are not included in the CRADLE dataset, we convert individual visits into natural language through textualization.

\noindent \textbf{Tasks.} On the MIMIC-III~\cite{johnson2016mimic} dataset, we perform phenotyping prediction, which involves predicting the presence of 25 care conditions in patients' next visits~\cite{harutyunyan2019multitask}, given their current ICU records. This can be useful for detecting morbidity, repurposing drugs, and diagnosis. On the CRADLE dataset, the task aims to determine if patients diagnosed with type 2 diabetes will experience cardiovascular disease (CVD) endpoints within a year of their diagnosis. CVD endpoint is defined by the presence of coronary heart disease (CHD), congestive heart failure (CHF), myocardial infarction (MI), or stroke. As CVD affects around 32\% of patients with diabetes~\cite{einarson2018prevalence}, it is essential to have a systematic CVD risk prediction.

\noindent \textbf{Metrics.} To account for the imbalanced label distribution in both MIMIC-III and CRADLE, we evaluate the performance using four metrics: Accuracy, AUROC, AUPR, and Macro-F1 score~\cite{choi2020learning, cai2022hypergraph}. We set the threshold for accuracy and F1 score to 0.5 for a fair comparison with the previous methods.

\noindent \textbf{Implementation Datails.} 
Our model is built with PyTorch. We use Adam as the optimizer with a learning rate of $1e^{-3}$. The weight decay is set to $1e^{-3}$.
We tune several key hyperparameters, including the hidden dimension $d$, the number of layers $L$ in the hypergraph neural network, and the dimension ratio between structural and semantical embeddings. Details are further discussed in Section~\ref{sec:hyperparam}.

\noindent \textbf{Baselines.} 
We compare \ours with various types of baselines, including
(1) \textit{Conventional ML Baselines. } Logistic Regression (LR), SVM, and MLP are selected as non-graph modeling baselines.
(2) \textit{GNN Baselines.} In graph-based methods, the graph is constructed based on pair-wise relations among medical codes: an edge is created between two codes if they co-occur in the same visit. We choose GCT~\citep{choi2020learning} and GAT~\citep{gat}.
(3) \textit{Hypergraph Modeling Baselines.} These baselines are tested using the same hypergraph structured as \ours but with various neural network architectures. We include HGNN~\citep{feng2019hypergraph}, HyperGCN~\citep{yadati2019hypergcn}, HCHA~\cite{bai2021hypergraph}, and HypEHR~\citep{xu2023hypergraph}.


\vspace{-0.5ex}
\begin{table*}[h]
\vskip -0.5em
\caption{Performance (100\%) on MIMIC-III and CRADLE compared with different baselines. The result is averaged over 5 runs. We use * to indicate statistically significant results ($p<0.05$). Bold and underlined indicate the best and second-runner results.}
\vskip -1.5em
\renewcommand\arraystretch{0.9}
\begin{center}
\resizebox{0.95\linewidth}{!}{
\begin{tabular}{lcccccccc}
\toprule
\multirow{2.5}{*}{\textbf{Model}} & \multicolumn{4}{c}{\textbf{MIMIC-III}}    &  \multicolumn{4}{c}{\textbf{CRADLE}}           \\ 
\cmidrule(lr){2-5} \cmidrule(lr){6-9}
&  ACC &  AUROC &  AUPR &  F1 & ACC &  AUROC &  AUPR &  F1\\
\midrule
LR & 68.66 ± 0.24 & 64.62 ± 0.25 & 45.63 ± 0.32 & 13.74 ± 0.40 & 76.22 ± 0.30 & 57.22 ± 0.28 & 25.99 ± 0.26 & 42.18 ± 0.35\\
SVM & 72.02 ± 0.12 & 55.10 ± 0.14 & 34.19 ± 0.17 & 32.35 ± 0.21 & 68.57 ± 0.13 & 53.57 ± 0.11 & 23.50 ± 0.15 & 52.34 ± 0.22\\
MLP & 70.73 ± 0.24 & 71.20 ± 0.22 & 52.14 ± 0.23 & 16.39 ± 0.30 & 77.02 ± 0.17 & 63.89 ± 0.18 & 33.28 ± 0.23 & 45.16 ± 0.26\\ \midrule
GCT & 76.58 ± 0.23 & 78.62 ± 0.21 & 63.99 ± 0.27 & 35.48 ± 0.34 & 77.26 ± 0.22 & 67.08 ± 0.19 & 35.90 ± 0.20 & 56.66 ± 0.25\\
GAT & 76.75 ± 0.26 & 78.89 ± 0.12 & 66.22 ± 0.29 & 34.88 ± 0.33 & 77.82 ± 0.20 & 66.55 ± 0.27 & 36.06 ± 0.18 & 56.43 ± 0.26\\
\midrule
HGNN & 77.93 ± 0.41 & 80.12 ± 0.30 & 68.38 ± 0.24 & 40.04 ± 0.35 & 76.77 ± 0.24 & 67.21 ± 0.25 & 37.93 ± 0.18 & 58.05 ± 0.23\\
HyperGCN & 78.01 ± 0.23 & 80.34 ± 0.15 & 67.68 ± 0.16 & 39.29 ± 0.20 & 78.18 ± 0.11 & 67.83 ± 0.18 & 38.28 ± 0.19 & {60.24 ± 0.21}\\
HCHA & 78.07 ± 0.28 & 80.42 ± 0.17 & 68.56 ± 0.15 & 37.78 ± 0.22 & 78.60 ± 0.15 & {68.05 ± 0.17} & 39.23 ± 0.13 & 59.26 ± 0.21\\
HypEHR & 79.07 ± 0.31 & 82.19 ± 0.13 & 71.08 ± 0.17 & 41.51 ± 0.25 & \underline{79.76 ± 0.18} & 70.07 ± 0.13 & 40.92 ± 0.12 & 61.23 ± 0.18 \\
\midrule
\rowcolor{teal!18} {\ours} &\textbf{80.17 ± 0.08}$^*$ & \textbf{83.54 ± 0.06}$^*$  & \textbf{72.50 ± 0.07}$^*$ & \textbf{46.26 ± 0.61}$^*$ & 78.87 ± 0.48 & \textbf{73.01 ± 0.06}$^*$ & \textbf{45.76 ± 0.13}$^*$ & \textbf{63.49 ± 0.49}$^*$ \\
\rowcolor{teal!7} {\ours} w/o Medical Concept Semantics  & 79.08 ± 0.18 & 82.37 ± 0.14$^*$ & 70.98 ± 0.26 & 41.83 ± 1.89 & \textbf{80.07 ± 0.38}$^*$ & \underline{72.49 ± 0.26}$^*$ & \underline{44.63 ± 0.24}$^*$ & 60.62 ± 1.53 \\
\rowcolor{teal!7} {\ours} w/o Clinical Note Semantics  & \underline{79.77 ± 0.33}$^*$ & \underline{83.14 ± 0.18}$^*$ & \underline{72.02 ± 0.32}$^*$ & \underline{45.69 ± 2.68}$^*$ & 75.39 ± 1.34 & 70.83 ± 0.62$^*$ & 43.90 ± 0.90$^*$ & \underline{63.19 ± 0.60}$^*$ \\
\midrule 
\bottomrule
\end{tabular}
}
\vskip -0.5em
\label{tab:main}

\end{center}
\end{table*}


\subsection{Experiment Results}
The results comparison on two EHR datasets of {\ours} and baselines are present in Table~\ref{tab:main}.
The results show that \ours achieves the best performance compared to all baselines on four metrics on the MIMIC-III dataset, particularly in a higher F1 score. On the CRADLE dataset, \ours shows a significant improvement over AUROC and AUPR. Since the datasets are unbalanced, the slight drop in accuracy with improvements in all other metrics may be due to an improvement in the class with a smaller sample size.

\vspace{-1.2ex}
\subsection{Ablation and Hyperparameter Studies}
\label{sec:ablation}
\noindent \textbf{Ablation Study of Two-level Semantics Fusion.} 
The last two rows of Table~\ref{tab:main} examine the effect of the two-level infusion strategy using ablation analysis. Our findings indicate that the removal of medical concept semantics consistently results in a significant decrease compared to the full \ours. This suggests that medical concept name semantics play a vital role in reasoning over structured data. In comparison, clinical note semantics appear to have less impact, possibly due to the challenge of constructing lengthy document representations from LLMs, particularly with the noisy nature of clinical notes. It is worth noting that for the CRADLE dataset, the clinical notes utilized are obtained from patient visit records, resulting in cleaner and mostly medical concept-based notes. Hence, the influence of removing hyperedge fusion is more pronounced in terms of AUROC and AUPR than with natural notes.

\noindent  \textbf{Effect of Hyperparameters.} 
\label{sec:hyperparam}
The influence of hidden dimension $d$ (24, 48, 72, 96) and the number of layers $L$ (1, 2, 3, 4) in the hypergraph model, and the dimension ratio (0.5. 0.67, 1, 1.5, 2) between structural and semantical embedding in Eq.~(\ref{eq:infuse_concept}) and ~(\ref{eq:infuse_note}) are investigated. Results are omitted here due to the space limit. 

\vspace{-1ex}
\section{Case Study}
\begin{figure}
	\centering
	\includegraphics[width=\linewidth]{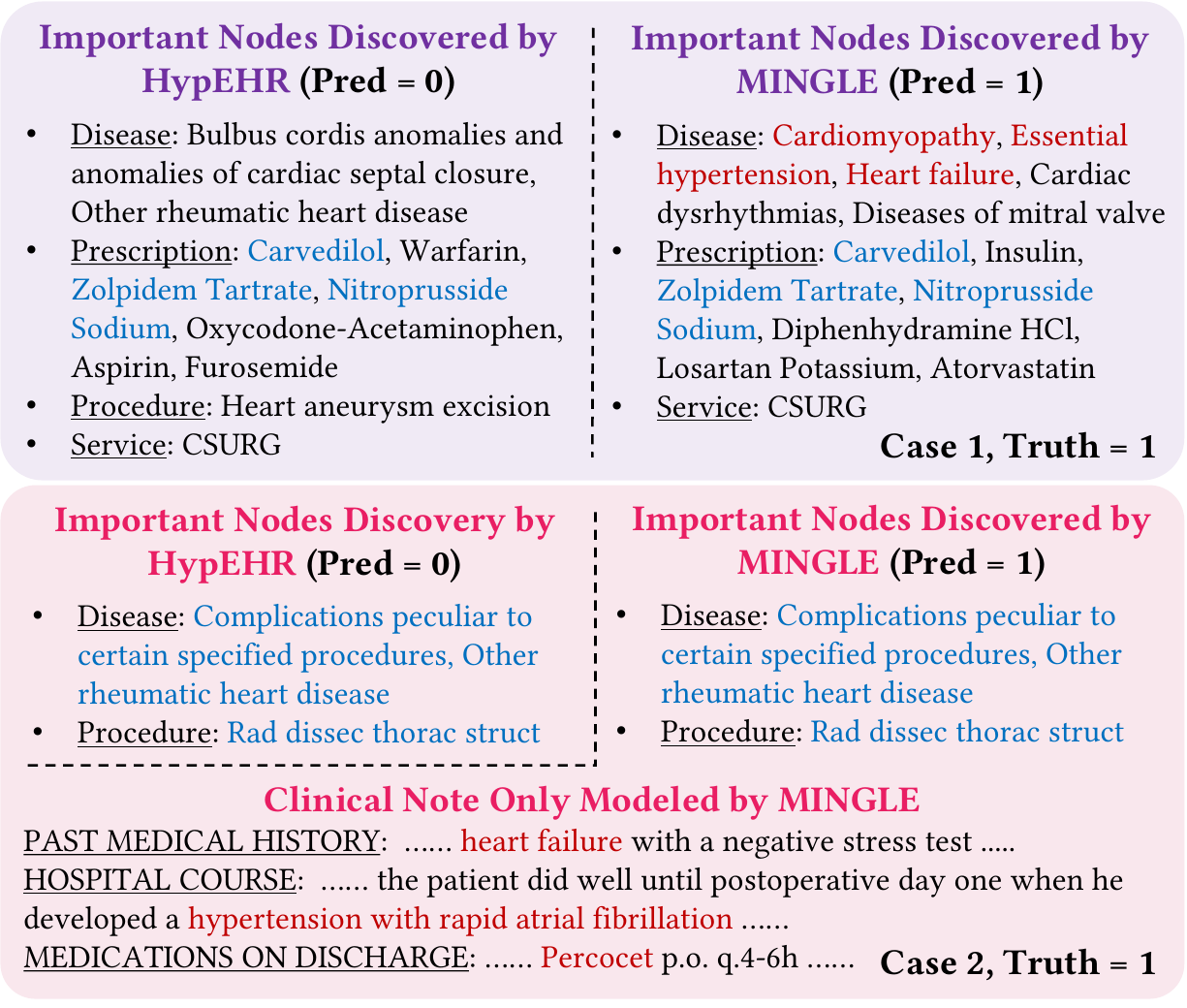}
 \vskip -0.5em
	\caption{Case studies on the important nodes highlighted by the hypergraph backbone model and \ours-- {\color{sblue}Blue} highlights the same important nodes discovered by both, while {\color{sred}Red} highlights the additional ones discovered by \ours. Case 2 includes the clinical notes only modeled by \ours, with {\color{sred}Red} highlighting the potentially important chunks.}
 \vskip -0.5em
\label{fig:case_study}
\vskip -0.5em
\end{figure}

We present two case studies on MIMIC-III \textit{Cardiac Dysrhythmias} phenotype prediction in Figure~\ref{fig:case_study} to demonstrate the difference in important medical node selection between the baseline and \ours based on attention weights in the self-attention mechanism. 

\noindent $\star$ \textbf{Case 1.} 
The differences and similarities between the important nodes discovered by the HypEHR and \ours models can help explain their differences in predicting \textit{Cardiac Dysrhythmias}.
The nodes identified by both models include \textit{Carvedilol}, a beta-blocker used for treating hypertension and heart failure and reducing the risk of arrhythmias, a significant indicator of cardiovascular disease and arrhythmia risk; \textit{Nitroprusside Sodium}, which is applied in acute heart failure and hypertensive emergencies; and \textit{Zolpidem Tartrate}, which can indicate insufficient rest and anxiety, and sleep quality can have a potential impact on cardiac health. All those nodes are closely related to \textit{Dysrhythmias}.
However, \ours also identified \textit{Heart Failure}, \textit{Cardiomyopathy}, and \textit{Cardiac Dysrhythmias}. These are diseases directly related to cardiac function. This suggests that the infusion of medical concept semantics in the \ours model introduced a deeper understanding of the clinical context, leading to an improved predictive ability because of the effective modeling of important nodes directly related to the prediction target. 

\noindent $\star$ \textbf{Case 2.} 
In the second case, the important node sets from HypEHR and our system are quite similar. However, clinical notes contain helpful information that can be used for better learning. For instance, the patient's \textit{past medical history} reveals that he had \textit{heart failure nearing the time of admission}. The \textit{hospital course} indicates that the patient developed \textit{rapid atrial fibrillation} and \textit{hypertension}. This indicates that the patient faced some issues related to their cardiac function after the surgery. 
Furthermore, \textit{Percocet}, a medication that contains oxycodone, can have \textit{side effects on the respiratory and cardiovascular systems}. 
\ours can combine the semantics of these clinical notes with the structural learning of EHR, leading to more comprehensive learning of patient profiles.





\vspace{-1.1ex}
\section{Conclusion and Discussion}
\vspace{-0.3ex}

Our \ours framework combines EHR clinical records and notes by infusing two-level semantics into hypergraph neural networks. Results show significant advantages of integrating medical concept semantics. In the future, we plan to investigate explicit extraction using LLMs and supplement multimodal EHR by aligning and fusing cross-modality data in both hard and soft ways.

\balance
\bibliographystyle{ACM-Reference-Format}
\bibliography{sample-base}

\end{document}